\documentclass[10pt,twocolumn,twoside]{IEEEtran}

\usepackage{cite}
\newcommand\citep{\cite}

\ifCLASSINFOpdf
   \usepackage[pdftex]{graphicx}
\else
   \usepackage[dvips]{graphicx}
\fi
\usepackage{epsfig}
\usepackage{multirow}
\usepackage[usenames, dvipsnames]{color}
\usepackage{subfig}
\usepackage{amsmath}
\usepackage{enumerate}
\usepackage{CJKutf8}
\usepackage{amsfonts}
\usepackage{comment}
\usepackage{algorithm}
\usepackage{algorithmic}
\usepackage{array}
\usepackage{hyperref}

\DeclareMathOperator*{\E}{\mathbb{E}}
\newcolumntype{L}[1]{>{\raggedright\arraybackslash}m{#1}}
\newcolumntype{N}{@{}m{0pt}@{}}

\begin{document}

\title{Improving Conditional Sequence Generative Adversarial Networks by Stepwise Evaluation} 
\author{Yi-Lin Tuan and Hung-Yi Lee\\
  Department of Electrical Engineering, National Taiwan University \\
\thanks{This work was financially supported by the Ministry of Science and Technology of Taiwan.}} 
\maketitle

\begin{abstract}
Sequence generative adversarial networks (SeqGAN) have been used to improve conditional sequence generation tasks, for example, chit-chat dialogue generation. To stabilize the training of SeqGAN, Monte Carlo tree search (MCTS) or reward at every generation step (REGS) is used to evaluate the goodness of a generated subsequence. MCTS is computationally intensive, but the performance of REGS is worse than MCTS. In this paper,  we propose stepwise GAN (StepGAN), in which the discriminator is modified to automatically assign scores quantifying the goodness of each subsequence at every generation step. StepGAN  has significantly less computational costs than MCTS. We demonstrate that StepGAN outperforms previous GAN-based methods on both  synthetic experiment and chit-chat dialogue generation.
\end{abstract}


\IEEEpeerreviewmaketitle

\section{Introduction}
Conditional sequence generation refers to tasks in which a response is generated according to an input.
Such tasks include machine translation, summarization, question answering and dialogue generation.
In these applications, the input messages and output responses usually have a {\bf one-to-many property}.
For example, in dialogue generation, given a message (\emph{``How was your day?''}), there are many acceptable responses (\emph{``It is good.''}, \emph{``Very bad.''}).
This property, mostly appears in dialogue generation, attributes to a conditional model to learn a distribution instead of a specific answer, thus preventing the model from generating high quality answers.

Generally, conditional sequence generation is learned using a sequence-to-sequence model (seq2seq) trained by minimizing the cross-entropy loss~\cite{vinyals2015neural}, a method often called maximum likelihood estimation (MLE)~\cite{vinyals2015neural, li2017adversarial}.
MLE achieved acceptable results in terms of coherence, but leads to three problems:
(1) \textbf{Exposure bias}~\cite{ranzato2015sequence}.
Targets and generated responses are respectively taken as the inputs during training stage and inference stage, a bias that causes accumulated errors in inference stage.
(2) \textbf{One directional KL divergence}~\cite{arjovsky2017towards}.
MLE only minimizes forward KL divergence and ignores the backward, causing the predicted distribution unbounded.
(3) \textbf{General Responses}. 
Previous work~\cite{li2016diversity} empirically found that MLE tends to produce general responses (e.g., \emph{``i don't know.''}), leading to dialogues with little information.
GANs have potential to solve the above three issues.

To tackle the problems of MLE, SeqGAN is recently proposed for chit-chat dialogue generation~\cite{yu2017seqgan, li2017adversarial}.
By using a discriminator to evaluate the difference between the model responses and ground truths, the problem of only one direction KL divergence is solved.
Additionally, exposure bias is eliminated using policy gradient~\cite{ranzato2015sequence} for optimization.
By observation SeqGAN can generate more creative sentences~\cite{yu2017seqgan, li2017adversarial}, however, its optimization remains inefficient.

SeqGAN uses 
policy gradient
to solve the intractable backpropagation issue, but a common problem in reinforcement learning appears:  sparse reward, that the non-zero reward is only observed at the last time step.
The primary disadvantage of sparse reward is making the training sample inefficient. 
The inefficiency slows down training because the generator has very little successful experience.
\textcolor{black}{Sparse reward causes another problem to chit-chat chatbot. In chatting,} an incorrect response and a correct one can share the same prefix.
For example, \emph{``I'm John.''} and \emph{``I'm sorry.''} have the same prefix \emph{``I'm''}. But for the input \emph{``What 's your name?''}, the first response is reasonable and the second one is weak.
\textcolor{black}{The same prefix then receives opposite feedbacks. The phenomenon continuously happens during training, making the training signals highly variant. The training is therefore unstable~\cite{yu2017seqgan,li2017adversarial,ranzato2015sequence,bahdanau2016actor}.}

To deal with sparse reward, the original SeqGAN is trained with a stepwise evaluation method -- Monte Carlo tree search (MCTS)~\cite{yu2017seqgan}.  
MCTS stabilizes the training, 
but it is computationally intractable when dealing with large dataset. 
\textcolor{black}{To meet the necessary of large dataset for chatbot, }reward at every generation step (REGS) is proposed to replace MCTS but with worse performance~\cite{li2017adversarial}.
According to previous attempts~\cite{yu2017seqgan,li2017adversarial,che2017maximum,fedus2018maskgan}, we know that stepwise evaluation can remarkably affect the results, but it has not been thoroughly explored.

We propose an alternative stepwise evaluation method to replace MCTS -- StepGAN. 
\textcolor{black}{The motivation is to based on theory use a discriminator to estimate immediate rewards without computing tree search.}
StepGAN needs only little modification on the discriminator, but has significantly less computational costs than MCTS.
To evaluate the effects of StepGAN, we first quantify GANs ability in learning less general responses, different direction of KL divergence, and coverage of many acceptable answers.
Then we check which stepwise evaluation method better raises the advantages of GAN by comparing different stepwise evaluation methods on a synthetic task and a real-world scenario -- chit-chat dialogue generation.
In synthetic task, we show that StepGAN outperforms previous work by learning more balanced KL divergence. In dialogue generation, we show that StepGAN can produce more less general responses with higher sentence quality.
Although GANs may not benefit a model in all aspects, we empirically demonstrate that StepGAN can utilize the advantages of adversarial learning in conditional sequence generation the most.
Moreover, it is worth stating that although we only apply StepGAN on dialogue generation in this paper, the proposed approach can be applied to any task that can be formulated as conditional sequence generation, for example, machine translation, summarization and video caption generation.


\textcolor{black}{The contribution of this paper can be summarized to three stages:
\begin{itemize}
    \item The experiments demonstrate that GANs can improve some aspects of conditional sequence generation, but may worsen some aspects such as coherence in dialogue generation. 
    \item The proposed StepGAN benefits GANs' advantages the most but has trade-offs on GANs' cons.
    \item The proposed StepGAN is significantly faster than MCTS, and shows comparable performance.
\end{itemize}
}

\section{Related Works}
Neural based conditional sequence generation is first successfully trained using seq2seq model~\cite{sutskever2014sequence, vinyals2015neural}. The model is initially learned by minimizing the cross-entropy between the true data distribution and the model distribution.
The method, often called MLE, suffers from exposure bias, so beam search, scheduled sampling~\cite{bengio2015scheduled} and REINFORCE~\cite{williams1992simple,ranzato2015sequence} are proposed to solve the problem.
On the other hand, MLE also leads to general responses, so mutual information~\cite{li2016diversity} and adversarial learning~\cite{li2017adversarial} are used to produce creative sentences.

To optimize a task-specific score (e.g., BLEU~\cite{papineni2002bleu}), REINFORCE~\cite{ranzato2015sequence}, a reinforcement learning~\cite{sutton1998reinforcement} based algorithm, is first presented to guide seq2seq models.
Further, MIXER~\cite{ranzato2015sequence} integrates MLE and REINFORCE to reduce the exploration space. 
Afterwards, other reinforcement learning methods are also proposed to improve the performance, including actor-critic~\cite{bahdanau2016actor}, off-policy learning~\cite{kandasamy2017batch} and deep reinforcement learning for multi-turns dialogue~\cite{li2016deep}.

Recently, BLEU score is verified to be weakly correlated with human prior knowledge~\cite{liu2016not}.
Therefore GANs, an algorithm that automatically assigns scores to sentences, are applied to sequence modeling.
The related work includes SeqGAN~\cite{yu2017seqgan}, MaliGAN~\cite{che2017maximum}, REGS~\cite{li2017adversarial}, WGAN-GP~\cite{press2017language,rajeswar2017adversarial}, TextGAN~\cite{zhang2017adversarial}, RankGAN~\cite{lin2017adversarial} and MaskGAN~\cite{fedus2018maskgan}.
To conquer the intractable backpropagation through discrete sequences, policy gradient~\cite{yu2017seqgan}, Gumbel-softmax~\cite{kusner2016gans}, soft-argmax~\cite{zhang2016generating} and Wasserstein distance~\cite{gulrajani2017improved} are applied. 
\textcolor{black}{Among them, policy gradient is the most widely used.
An early version of using policy gradient on sequence modeling is REINFORCE.
As researchers have found that REINFORCE would cause high variance during training, MCTS~\cite{yu2017seqgan,li2017adversarial}, a stepwise evaluation method, is used to tabilize the training, however, it suffers from high computational costs.}
\textcolor{black}{To optimize the computation, several alternatives have been proposed to evaluate every subsequences~\cite{li2017adversarial, fedus2018maskgan}, but their performances are weaker than MCTS~\cite{li2017adversarial}.}

GANs have been applied to several tasks that can be formulated as conditional sequence generation. 
For dialogue generation, clear performance improvements on multiple metrics have been observed with SeqGAN using MCTS and REGS~\cite{ li2017adversarial}.
GANs improve machine translation models with different network architectures including RNNSearch~\cite{MTGAN_arXiv17,MTGAN_NAACL18} and Transformer~\cite{ MTGAN_NAACL18}.
For abstractive summarization, GANs are able to generate more abstractive, readable and diverse summaries than conventional approaches~\cite{Summarization_AAAI18}.
Image captioning models trained with GANs produce captions which are diverse and match the statistics of human generated captions significantly better than the baseline models~\cite{caption_ICCV17, caption_arXiv18, caption_arXiv17}.

We observe that there is still no research on comparing stepwise evaluation methods, even though many previous work~\cite{yu2017seqgan,li2017adversarial,che2017maximum,fedus2018maskgan} has noted the importance.
Therefore in this paper we study the effectiveness of stepwise evaluation methods on conditional sequence generation, and propose StepGAN which preserves the property of MCTS but with computational efficacy.
This is the first paper comparing the stepwise evaluation methods, and a StepGAN is proposed to replace them as a better training method for sequence GANs.

\begin{figure*}[t]
\centering
    \includegraphics[width=\textwidth]{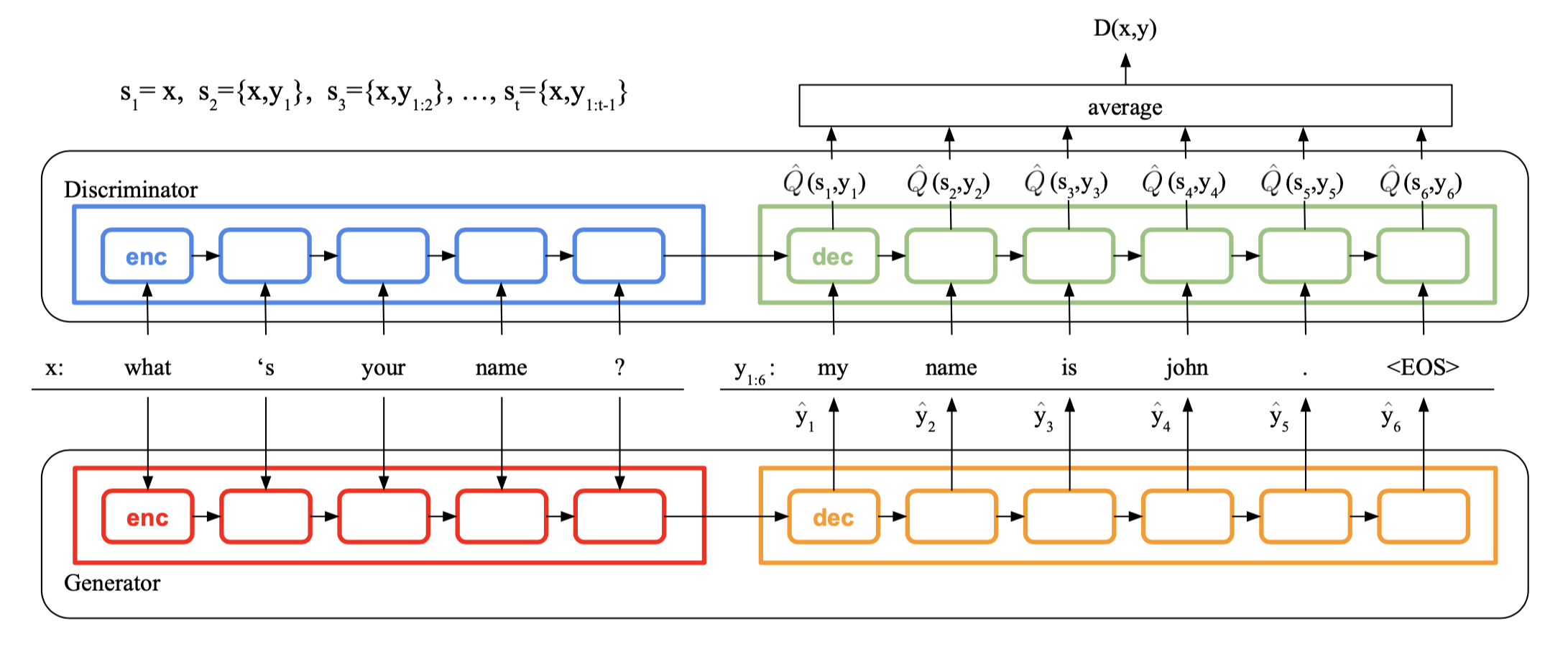}
    \caption[font=small]{The illustration of StepGAN. The generator's encoder (\textcolor{red}{red}) reads in sentence $x$, and then the generator's decoder (\textcolor{BurntOrange}{orange}) produces output sentence $\hat{y}$. The discriminator (\textcolor{blue}{blue} and \textcolor{LimeGreen}{green}) either reads in the generated sentence pair $(x,\hat{y})$ or the target sentence pair $(x, y^*)$, and then outputs estimated Q-values $\hat{Q}(s_t,\hat{y}_{t})$ (or $\hat{Q}(s_t,{y}^*_{t})$) at each time step $t$. The average of estimated Q-values at every time step $t$ is taken as the final discriminator score $D(x,\hat{y})$ (or $D(x,y^*)$).}
\label{fig:stepGAN}
\end{figure*}

\section{Conditional Sequence Generation}
When a model aims to produce an output sequence $y$ given an input $x$, it is called conditional sequence generative model.
For example, in chit-chat dialogue generation,
both $x$ and $y$ are word sequences defined as $x=\{x_1, x_2, ..., x_t, ..., x_N\}$ and $y=\{y_1, y_2, ..., y_t, ..., y_M\}$, where $x_t$ and $y_t$ are words in a vocabulary, and $N$ and $M$ are respectively the lengths of the input and output sequences. 

\subsection{Maximum Likelihood Estimation}
\label{sec:seq2seq}
The conditional sequence generator $G$ used in this paper is a recurrent neural network (RNN) based seq2seq model, consisting of an encoder and a decoder \textcolor{black}{with embedding layers}. 
The encoder reads in an input sentence one token $x_t$ at a time.
\textcolor{black}{The decoder aims to generate an output sentence $\hat{y}$ as close as possible to the target $y^*$.}

\textcolor{black}{During inference, after the encoder reads the whole $x$, the decoder generates the first word distribution $P_G(y_1|x)$.}
The first word $\hat{y}_1$ is obtained by sampling or \textit{argmax} from $P_G(y_1|x)$.
The decoder then takes $\hat{y}_1$ as the next input token, and generates the next word distribution $P_G(y_2|x,\hat{y}_1)$.
When the token representing the end of the sentence is predicted, the generation process stops, and the output sequence $\hat{y}$ is completely generated.
Because sampling is used in the generation process, the generator can be considered as a distribution of $y$ given $x$:  $P_G(y|x)=\prod_{t=1}^{M}P_G(y_t|x,y_{1:t-1})$, where $y_{1:t-1}$ represents the first $(t-1)$-th tokens in sequence $y$, or $y_{1:t-1}=\{y_1,y_2,...,y_{t-1}\}$.
For instance if \textit{argmax} is used in the decoder, we can consider $P_G$ as a Kronecker delta function (or a very sharp distribution) that the probabilities are zeros everywhere except for the chosen sample.

The model is trained to minimize the cross-entropy of $\hat{y}$ and the ground-truth sequence $y^*$.
This method is often called \textit{maximum likelihood estimation} (MLE)~\cite{vinyals2015neural}.
However, the inconsistency of decoder inputs between training stage and inference stage leads to \textit{exposure bias}~\cite{ranzato2015sequence}, whereby the generator would accumulate errors in inference stage. 

\subsection{Generative Adversarial Network (GAN)}
\label{sec:gan}
A GAN, consists of a generator $G$ and a discriminator $D$, for sequence learning is often called sequence GAN.
Given an input sequence $x$, the discriminator learns to maximize score $D(x,y^*)$ and minimize score $D(x,\hat{y})$, while the generator learns to produce response $\hat{y}$ to maximize $D(x,\hat{y})$. 
\begin{equation}
  \label{eq:gan}
  \begin{split}
    \underset{G}{\min} \underset{D}{\max} & \E_{ (x,y^*) \sim P_R(x,y) }[\log D(x,y^*)] \\
    & + \E_{x \sim P_R(x), \hat{y} \sim P_G(y|x)}[\log(1-D(x,\hat{y}))],
  \end{split}
\end{equation}
where $P_R(x)$ and $P_R(x,y)$ are the probability distribution of $x$ and joint probability distribution of $(x,y)$ from the training data.


To overcome the intractable gradients due to the discrete nature of language, previous work formulates sequence generation as a Markov decision process and trains a GAN using {\it policy gradient}~\cite{yu2017seqgan}.
The basic idea is to take the policy as the parameters of a generator $P_G$,
the state at time step $t$ as $\{x, \hat{y}_{1:t-1}\}$,
the action taken at time step $t$ as $\hat{y}_t$.
Specifically, the rewards except for the last time step are set to zero, and the final step reward is set to $D(x,y)$.
The intuition of training sequence GAN using policy gradient is that the generator maximizes the probabilities of words with higher expected $D(x,y)$ and minimizes others.
The gradient of this method can be approximated as follows.
\begin{equation}
    \nabla_G = \sum_{t=1}^M D(x,\hat{y}) \nabla \log P_G(\hat{y}_t|x,\hat{y}_{1:t-1})
\label{eq:pg}
\end{equation}
In Equation~(\ref{eq:pg}), the gradients of the log probabilities are weighted by the same weight $D(x,\hat{y})$.

Compared to MLE, a generator can minimize any loss defined by a discriminator instead of one direction KL divergence.
Moreover, policy gradient adopts the same $\hat{y}_{1:t-1}$ in inference and training stages, thus solving the problem of exposure bias.

\section{Stepwise Evaluation Approach \\ for Sequential GANs}

To improve sequence GAN by tackling both sparse reward and training instability, we propose a new stepwise evaluation method called {\bf StepGAN}.
Borrowing the idea of Q-learning, StepGAN automatically estimates state-action values with very light computational costs. 
The framework of StepGAN is depicted in Figure~\ref{fig:stepGAN}.

\begin{algorithm}[!htbp]
  \caption{StepGAN}
  \label{alg:algorithm1}
  \begin{algorithmic}[1]
 \STATE Initialize generator $G$\
 \STATE Pretrain generator $G$ using MLE
 \STATE Initialize discriminator $D$, value network $V$
 \FOR{number of training iterations}
     \FOR{$i=1$ to D-iterations}
     \STATE  Sample $(x,y^*)$ from real data
     \STATE  Sample $\hat{y} \sim P_G(y|x)$
     \STATE Update D using equation~(\ref{eq:step-D})
     \ENDFOR
     \STATE  Update $V$ by
     \begin{equation}\nonumber
         \underset{V}{\min}\left\|V(s_t)-Q(s_t,\hat{y}_{t})\right\|_2
     \end{equation}
     \FOR{$i=1$ to G-iterations}
     \STATE  Sample $x$ from real data
     \STATE Sample $\hat{y} \sim P_G(y|x)$
     \STATE Update $G$ using equation~(\ref{eq:step-G})
      \ENDFOR
 \ENDFOR
\end{algorithmic}
\end{algorithm}

\subsection{State-action Value}

In Figure~\ref{fig:stepGAN}, each $Q(s_t, y_{t})$ is a state-action value when taking action $y_t$ at state $s_t=\{x, y_{1:t-1}\}$.
This value, also a general form of Q-value, is often used in reinforcement learning to stabilize the training~\cite{sutton1998reinforcement, mnih2013playing, mnih2015human, hausknecht2015deep}.
The definition of state-action value in conditional sequence GAN is as below.
\begin{equation}
\label{eq:Q}
    Q(s_t,\hat{y}_{t}) = \underset{z\sim P_G(.|x,\hat{y}_{1:t})}{\E} [D(x,\{\hat{y}_{1:t},z\})].
\end{equation}
where $z$ is a sequence of words generated by the current generator given input $x$ and generated prefix $\hat{y}_{1:t}$.
Accordingly, the state-action value $Q(s_t,\hat{y}_{t})$ is the expected return of all the responses sharing the same prefix $\hat{y}_{1:t}$. 
The input of the Q-function are $s_t=\{x, y_{1:t-1}\}$ and $y_t$, which are discrete tokens. 
Because the Q-values are  the outputs of the discriminator, which has embedding layer, we can consider that the tokens are transformed into embedded forms within the Q-function.

To stabilize \textcolor{black}{the REINFORCE algorithm for GAN}, the term $D(x,\hat{y})$ in Equation~(\ref{eq:pg}) is replaced with $Q(s_t,\hat{y}_{t})$, and the equation is modified as:
\begin{equation}
    \nabla_G = \sum_{t=1}^M Q(s_t,\hat{y}_{t}) \nabla \log P_G(\hat{y}_t|x,\hat{y}_{1:t-1})
\label{eq:pg_q}
\end{equation}
In Equation~(\ref{eq:pg_q}), each generation step is weighted by a step-dependent value, $Q(s_t,\hat{y}_{t})$. 
However, computing $Q(s_t,\hat{y}_{t})$ by Equation~(\ref{eq:Q}) is intractable. 
We propose an efficient approach to estimate $Q(s_t,\hat{y}_{t})$ for generator while training the discriminator in StepGAN.



\subsection{Discriminator for StepGAN}\label{sec:step-D}
In this paper, the discriminator is a seq2seq model that takes $x$ as encoder inputs, and $y$ (either $\hat{y}$ or $y^*$) as decoder inputs.
Although in Figure~\ref{fig:stepGAN}, the discriminator is only a native seq2seq model, any state-of-the-art seq2seq model, for example, attention-based model~\cite{ShowAttTell}, CNN-based model~\cite{CNN_seq2seq}, Transformer~\cite{Transformer}, etc., can be used here.
As regular GANs, the discriminator is trained to maximize $D(x,y^*)$ of ground-truth examples and minimize the $D(x,\hat{y})$ of generated response as below.
\begin{equation}
\label{eq:step-D}
\begin{split}
&\underset{D}{\max} \E_{ (x,y^*) \sim P_R(x,y) }[\log D(x,y^*)] \\
&+ \E_{x \sim P_R(x), \hat{y} \sim P_G(y|x)}[\log(1-D(x,\hat{y}))]
\end{split}
\end{equation}

However, different from all the previous GAN-based sequence generation approaches, the definition of $D(x,y)$ in StepGAN is designed to estimate the  $Q(s_t,y_{t})$.
Instead of generating a scalar as the final discriminator score $D(x,y)$~\cite{yu2017seqgan}, we do some modification to the discriminator to automatically obtain the estimation of $Q(s_t,y_{t})$. 
As shown in the  upper block of Figure~\ref{fig:stepGAN}, after reading in the input sentence $x$ and part of the response sequence $y_{1:t}$, discriminator generates a scalar $\hat{Q}(s_t,y_{t})$. 

The discriminator score $D(x,y)$ is the average of all the scalars $\hat{Q}(s_t,y_{t})$ throughout the generated sequence length $M$.
\begin{equation}
\label{eq:step-D-avg}
D(x,y) =     \frac{1}{M}\Sigma_{t=1}^{M} \hat{Q}(s_t,y_{t}),
\end{equation}
The term $\hat{Q}(s_t,y_{t})$ in Equation~(\ref{eq:step-D-avg}) directly matches the role of $Q(s_t,y_{t})$. 
Therefore, we take the scalars $\hat{Q}(s_t,y_{t})$ as the approximation of state-action values $Q(s_t,y_{t})$ to train the generator in  Section~\ref{sec:stepGAN-G}.
The theory behind the above approximation will be clear in Section~\ref{sec:stepGAN-th}.

\subsection{Generator for StepGAN}\label{sec:stepGAN-G}
The generator reads in the input sentence $x$ and then predicts one token $\hat{y}_t$ at a time based on the generated word sequence $\hat{y}_{1:t-1}$.
For stepwise evaluation, we update generator $G$ by replacing $Q(s_t,\hat{y}_{t})$ with $\hat{Q}(s_t,\hat{y}_{t})$ in Equation~(\ref{eq:pg_q}).

As previous works~\cite{ranzato2015sequence,li2017adversarial}, we train a value network $V$ to generate the value baseline for stabilizing policy gradient. 
The value network has the same structure as discriminator.
It predicts the expected value $V(s_t)$.
The value network is trained to approximate the predicted $\hat{Q}(s_t,y_{t})$ for every previous states $s_t$. That is, $V(s_t) = \underset{y_t}{\E}[\hat{Q}(s_t,y_{t})]$.

With the value network, we train the generator as below.
\begin{equation}
\label{eq:step-G}
	\begin{split}
		\nabla_G = & \sum_{t=1}^M \alpha_t(\hat{Q}(s_t,\hat{y}_{t})-V(s_t)) \\
        & \nabla \log P_G(\hat{y}_t|x,\hat{y}_{1:t-1}).
    \end{split}
\end{equation}
where $\alpha_t$ is a weighting coefficient that is related to time $t$.
In uniform case, $\alpha_t$ equals to $1$ for all time $t$. We also test increasing and decaying cases in the following experiments.

In Figure~\ref{fig:stepGAN}, the generator is a seq2seq model, but the proposed approach is independent to the network architecture of the generator.
The complete training process of StepGAN is illustrated in  Algorithm 1.

\begin{table*}[t]
    \centering
    \resizebox{\textwidth}{!}{\begin{tabular}{l|l|lN}
         \bf Methods & \bf computation of D(x,y) & \bf optimization of G\\\hline
         SeqGAN & $D(x,y) = \hat{Q}(x,y_{1:M})$ & $\nabla_G = \sum_{t=1}^M D(x,y) \nabla \log P_G(y_t|x,y_{1:t-1})$ &\\[5pt]
         \hline
         REGS & $D(x,y) \sim \{\hat{Q}(s_t,y_{t})\}_{t=1}^{M}$ & $\nabla_G = \sum_{t=1}^M \hat{Q}(s_t,y_{t}) \nabla \log P_G(y_t|x,y_{1:t-1})$ & \\[5pt]
         \hline
         MCTS & $D(x,y) = \hat{Q}(x,y_{1:M})$ & $\nabla_G = \sum_{t=1}^M Q^*(s_t,y_{t}) \nabla \log P_G(y_t|x,y_{1:t-1})$,\\
         & &  where $Q^*(s_t,\hat{y}_{t}) = \frac{1}{I} \sum_{i=1, z_i\sim P_G(.|x,y_{1:t})}^{I} [D(x,\{y_{1:t},z_i\})].$ & \\[5pt]
         \hline
         MaskGAN & $D(x,y) = \{D(x,y_{1:t})\}_{t=1}^M = \{\hat{Q}(s_t,y_{t})\}_{t=1}^M$ & $\nabla_G = \sum_{t=1}^M Q^*(s_t,y_{t}) \nabla \log P_G(y_t|x,y_{1:t-1})$, & \\[5pt]
         & &  where $Q^*(s_t,\hat{y}_{t}) = \sum_{\tau=t}^{M} [D(x,y_{1:\tau})].$ & \\[5pt]
         \hline
         StepGAN & $D(x,y) = \frac{1}{M} \sum_{t=1}^{M} \{\hat{Q}(s_t,y_{t})\}_{t=1}^{M}$ & $\nabla_G = \sum_{t=1}^M \hat{Q}(s_t,y_{t}) \nabla \log P_G(y_t|x,y_{1:t-1})$ & \\[5pt]
    \end{tabular}}
    \caption{The equations for different stepwise evaluation methods. The details are explained in section~\ref{sec:others-eqs}.}
    \label{tab:others-eqs}
\end{table*}

\subsection{Theory behind StepGAN} \label{sec:stepGAN-th}

StepGAN is based on a direct linking between $D(x,y)$ and $Q(s_t,y_{t})$ as below. 
\begin{equation}
\label{eq:stepGAN}
    \underset{\hat{y} \sim P_G(y|x)} {\E}[D(x,\hat{y})]
    = \frac{1}{M} \underset{\hat{y} \sim P_G(y|x)}{\E}[\sum_{t=1}^{M}Q(s_t,\hat{y}_{t})].
\end{equation}
The equation above suggests that we can decompose $D(x,y)$ into $M$ different scores for all the time steps $t$ and take the decomposed scores as $Q(s_t,y_{t})$. 
Because of Equation~(\ref{eq:stepGAN}), we estimate state-action values $Q(s_t,y_{t})$ by adding a simple component to the architecture of discriminator as shown in (\ref{eq:step-D-avg}).

We can derive Equation~(\ref{eq:stepGAN}) by following:
\begin{equation}
\begin{split}
    & \underset{\hat{y}_{1:M} \sim P_G(y|x)}{\E}
    \Big[\sum_{t=1}^{M} Q(s_t,\hat{y}_{t}) \Big] \\
    & = \underset{\hat{y}_{1:M} \sim P_G(y|x)}{\E}
    \Big[\sum_{t=1}^{M} \underset{z\sim P_G(.|x, \hat{y}_{1:t})}{\E}
    [r_{ter}(x, \hat{y}_{1:t}, z)]\Big], \\
\end{split}
\end{equation}
where $z$ is sampled from $P_G$ by given input $x$ and partial prefix $\hat{y}_{1:t}$, and $r_{ter}$ is the reward of the whole sequence $\{x, \hat{y}_{1:t}, z\}$. Based on GAN, the $r_{ter}$ is estimated by discriminator score $D$, so the equation can be rewritten as:
\begin{equation}
\begin{split}
&\underset{\hat{y}_{1:M} \sim P_G(y|x)}{\E}
    \Big[\sum_{t=1}^{M} \underset{z\sim P_G(.|x, \hat{y}_{1:t})}{\E}
    [r_{ter}(x, \hat{y}_{1:t}, z)]\Big] \\
    & = \underset{\hat{y}_{1:M} \sim P_G(y|x)}{\E}
    \Big[\sum_{t=1}^{M} \underset{z\sim P_G(.|x, \hat{y}_{1:t})}{\E}
    [D(x, \{\hat{y}_{1:t}, z\})]\Big] \\
    & = \sum_{t=1}^{M} \underset{\hat{y}_{1:M} \sim P_G(y|x)}{\E}
    \Big[\underset{z\sim P_G(.|x, \hat{y}_{1:t})}{\E}
    [D(x, \{\hat{y}_{1:t}, z\})]\Big]. 
\end{split}
\end{equation}
The two expectation $\E$ can be combined:
\begin{equation}
\begin{split}
&\sum_{t=1}^{M} \underset{\hat{y}_{1:M} \sim P_G(y|x)}{\E}
    \Big[\underset{z\sim P_G(.|x, \hat{y}_{1:t})}{\E}
    [D(x, \{\hat{y}_{1:t}, z\})]\Big] \\
    & = \sum_{t=1}^{M} \underset{\hat{y}_{1:M} \sim P_G(y|x)}{\E}
    \big[D(x, \hat{y}_{1:M})\big] \\
    & = M \cdot \underset{\hat{y}_{1:M} \sim P_G(y|x)}{\E}
    \big[D(x, \hat{y}_{1:M})\big]. 
\end{split}
\end{equation}
Then we divide the above equations by $M$ and obtain the equation below:
\begin{equation}
\label{eq:deriv}
\underset{\hat{y} \sim P_G(y|x)} {\E}[D(x,\hat{y})] = \frac{1}{M} \underset{\hat{y} \sim P_G(y|x)}{\E}[\sum_{t=1}^{M}Q(s_t,\hat{y}_{t})].
\end{equation}
The same equation can also be derived by substituting $P_G$ with $P_R$, and substituting $\hat{y}$ with $y^*$. Equation~(\ref{eq:step-D-avg}) is therefore a sample estimation of Equation~(\ref{eq:deriv}).

\begin{table*}
\begin{center}
\resizebox{.75\textwidth}{!}{\begin{tabular}{|l|ccc|cc|c|}
\hline  & \bf Prec & \bf SampP & \bf SampR & \bf FKLD & \bf IKLD & \bf FKLD+IKLD\\ \hline
MLE & 87.07 & 71.92 & 69.43
    & 0.6198 & 6.663 & 7.283\\
SeqGAN & 88.01 & 71.97 & \bf 69.85
    & 0.5969 & 6.635 & 7.232\\
REGS 
    & 88.26 & 72.37 & 68.43
    & 0.6729 & 6.601 & 7.274\\
MCTS
    & 88.55 & 72.16 & 68.87
    & 0.6559 & 6.614 & 7.270\\
MaskGAN 
    & 90.75 & 75.54 & 61.32
    & 1.092 & 6.120 & 7.212\\
StepGAN 
    & 90.94 & 72.70 & 69.47
    & 0.6363 & 6.484 & 7.121\\
StepGAN-W 
    & \bf 93.04 & \bf 73.83 & 69.67
    & 0.6904 & 6.282 & 6.973\\
\hline
\end{tabular}}
\end{center}
\caption{\label{tab:pretrain-count} Results of synthetic experiment - counting. }
\end{table*}

\section{Experiments: Baseline Approaches}
\label{sec:others-eqs}
The following stepwise evaluation methods are used as the baselines in the experiments. The equations are listed in Table~\ref{tab:others-eqs}.

\textbf{SeqGAN~\cite{yu2017seqgan}.} The discriminator predicts a final score $D(x,y)$ after reading the entire pair of sentences $(x,y)$. The generator adopts 1-sample estimation, using $D(x,y)$ as the multipliers for every time steps $t$, for policy gradient updates.

\textbf{Reward at Every Generation Step (REGS)~\cite{li2017adversarial}.} The discriminator optimizes a randomly selected score $D(x,y) \sim \{\hat{Q}(s_t,y_{t})\}_{t=1}^M$. 
Every scores $\{\hat{Q}(s_t,y_{t})\}_{t=1}^M$ are taken as the multipliers for policy gradient.

\textbf{Monte-Carlo tree search (MCTS)~\cite{yu2017seqgan, li2017adversarial, che2017maximum}.} 
MCTS computes the estimated state-action value $Q^*(s_t, y_t)$.
Given input sentence and fixed prefix $(x,y_{1:t})$, we roll out $I$ suffixes $y_{t+1:M_i}^i$ using the generator $G$, where $i$ is the label of suffix from $1$ to $I$, and each suffix has different lengths $M_i$.
Here $\{y_{1:t},y_{t+1:M_i}^i\}$ forms a full response whose first tokens are $y_{1:t}$, and the rest tokens are $y_{t+1:M_i}^i$.
We average $D(x,y)$ of the $I$ responses obtained by roll-out, $\frac{1}{I} \Sigma_{i=1}^I D(x, \{y_{1:t},y_{t+1:M_i}^i\}) $, as the approximated state-action value $Q^*(s_t, y_{t})$.
Monte-Carlo search would have high precision if $I$ is very large but with very high computational costs.

\textbf{MaskGAN~\cite{fedus2018maskgan}.} The discriminator optimizes all scores $D(x,y_{1:t}) = \hat{Q}(s_t,y_{t})$ for every time step $t$; the generator takes each score $D(x,y_{1:t})$ as received reward, and estimates state-action value by summing future received rewards: $Q^*(s_t,y_{t})=\sum_{\tau=t}^M [D(x,y_{1:\tau})]$. 



\textcolor{black}{The proposed approach differs from prior work  in two aspects.
First,  MCTS stabilizes the training of SeqGAN with lots of extra computation, but StepGAN stabilizes the training with nearly zero extras.
Second, while REGS and MaskGAN predict state-action values by heuristic, while StepGAN theoretically approximates a discriminator score as the average of correspondent state-action values. }

\section{Experiments: Synthetic Experiment}

\begin{table*}[t]
\begin{center}
\resizebox{.75\textwidth}{!}{\begin{tabular}{|l|c|cc|c|c|c|}\hline
 &\bf BLEU & \bf CoHS~(\%) &\bf SHS~(\%) &\bf LEN &\bf GErr &\bf General\\\hline
{\bf MLE} & 0.2580
		  &\bf 50.52
          & 42.6 & 5.765 & 32 & 115688\\
{\bf SeqGAN} & 0.2615
		  & 46.25
          & 56.4 & 6.525 & 38 & 92273\\
{\bf REGS} & 0.2540
		  & 48.43
          & 59.5 & 7.000 & 124 & 107719\\
{\bf StepGAN-W} & 0.2394
		  & 44.96
          &\bf 60.4 &\bf 7.335 &\bf 24 &\bf 51318\\\hline 
\end{tabular}}
\caption{The results of chit-chat dialogue generation. CoHS~(\%) is coherence human score; SHS~(\%) is semantics human score. To make sure the reliability of human scores, we have measured their intraclass correlation coefficient (ICC)~\cite{fleiss1973equivalence,bartko1966intraclass}, and they show substantially consistent.}
\label{tab:main-dialog}
\end{center}
\end{table*}

\label{sec:syn}

To compare their performances accurately, we conducted experiments on synthetic data. This also help us understand how the StepGAN works.
The source code is available at \href{https://github.com/Pascalson/Conditional-Seq-GANs}{\texttt{https://github.com/Pascalson/Conditional-\\Seq-GANs}}.

\subsection{Task Description}
A counting task is designed to capture the one-to-many property of conditional sequence generation tasks.
In a counting task, given an input sequence $x=\{x_1, x_2, ..., x_t, ..., x_N\}$, a correct output sequence $y=\{y_1, y_2, y_3\}$ obeys the following rules.
\begin{equation}
    \left\{
    \begin{array}{l}
        k \sim \{1, 2, ..., N\} \\
        y_1 = k-1 \\
        y_2 = x_k \\
        y_3 = N-k
    \end{array}
    \right .
\end{equation}
where each token $x_t$ is a digit selected from $0$ to $9$, and number $N$ is the length of input.

Based on the above setup, given the same input sequence, several different output sequences are correct.
For example, given an input sequence ${<1,8,3>}$, the possible answers are ${<0,1,2>}$, ${<1,8,1>}$ and ${<2,3,0>}$. 

We generated 100,000 training examples, 10,000 validation examples and 10,000 testing examples according to the counting rule.
\textcolor{black}{The maximum length of the input sequence was set to 10 ($N\leq10$)}\textcolor{black}{, and the average number of possible answers is 4.97.}
We evaluated different sequence generation approaches based on GANs.
All the GANs were trained upon a pretrained model trained by MLE until converged.
All the GANs were trained with $64$ minibatch size for $5,000$ iterations.

\subsection{Evaluation Metrics} 
In the synthetic experiments, because we know all the possible answers to an input sentence, we can compute the \emph{precision} and \emph{recall} rates.
Given each input $x$, we used the generator $G$ to generate one sample by \texttt{argmax} and 100 samples by sampling from the probability distribution of \texttt{softmax} outputs. 
In Table~\ref{tab:pretrain-count} three metrics are evaluated: (1) \textbf{Prec}, the precision of argmax outputs. It is not possible to evaluate recall for argmax because there is only one sample. 
(2) \textbf{SampP}, the precision of softmax outputs. (3) \textbf{SampR}, the recall of softmax outputs.
They are intuitive indexes to test the performance of the models.

To have more insights of the models, we analyzed generator by evaluating both the \emph{forward KL divergence (FKLD)} and \emph{inverse KL divergence (IKLD)} between the conditional distribution of the generator, $P_G(y|x)$, and the true distribution based on the counting rule, $P_R(y|x)$.
The definition of FKLD and IKLD are as below.
\begin{equation}
    \label{eq:KLD}
    \begin{split}
        FKLD& = \int_{x}\int_{y} P_R(y|x)\log \frac{P_R(y|x)}{P_G(y|x)}\\
        IKLD& = \int_{x}\int_{y} P_G(y|x)\log \frac{P_G(y|x)}{P_R(y|x)}\\
    \end{split}
\end{equation}

All the above metrics are the benefits of using synthetic tasks. In real applications like dialogue generation, it is not possible to list all the correct responses, so \emph{precision} and \emph{recall} cannot be accurately measured. Additionally, the computations of \emph{FKLD} and \emph{IKLD} in most real applications are intractable; their computations in the synthetic task is tractable due to the finite number of answers.

\subsection{Results}
The results are shown in Table~\ref{tab:pretrain-count}.
We compared StepGAN with other stepwise evaluation methods including REGS, MCTS, MaskGAN. 
In addition, we performed weighted stepGANs (stepGAN-W) that we gave decaying weights for the policy gradient.
The weights in Equation~(\ref{eq:step-G}) were set as $\alpha_t = M-t$, where $M$ was the length of the generated outputs.
We had tested both increasing and decaying cases, and found that increasing weights did not improve the training, so the results of  increasing weights were not shown here.

{\bf Precision and Recall.} Compared to MLE, all GANs improve the precision of argmax outputs by \textbf{Prec} in Table~\ref{tab:pretrain-count}.
Specifically, SeqGAN improves the model by $0.94$; previous stepwise evaluation methods improve the model by $1.19\sim 3.68$; our proposed approaches StepGAN and StepGAN-W improve the model by $3.87$ and $5.97$ respectively.

For sampled softmax outputs, in Table~\ref{tab:pretrain-count} the GANs have better precision (the column labeled \textbf{SampP}) and weaker recall (the column labeled \textbf{SampR}). Especially, the MaskGAN increases the precision the most but significantly drops the recall. This indicates that MaskGAN overfits to a smaller portion of possible answers.

\begin{figure}[t!]
\centering
    \includegraphics[width=\linewidth]{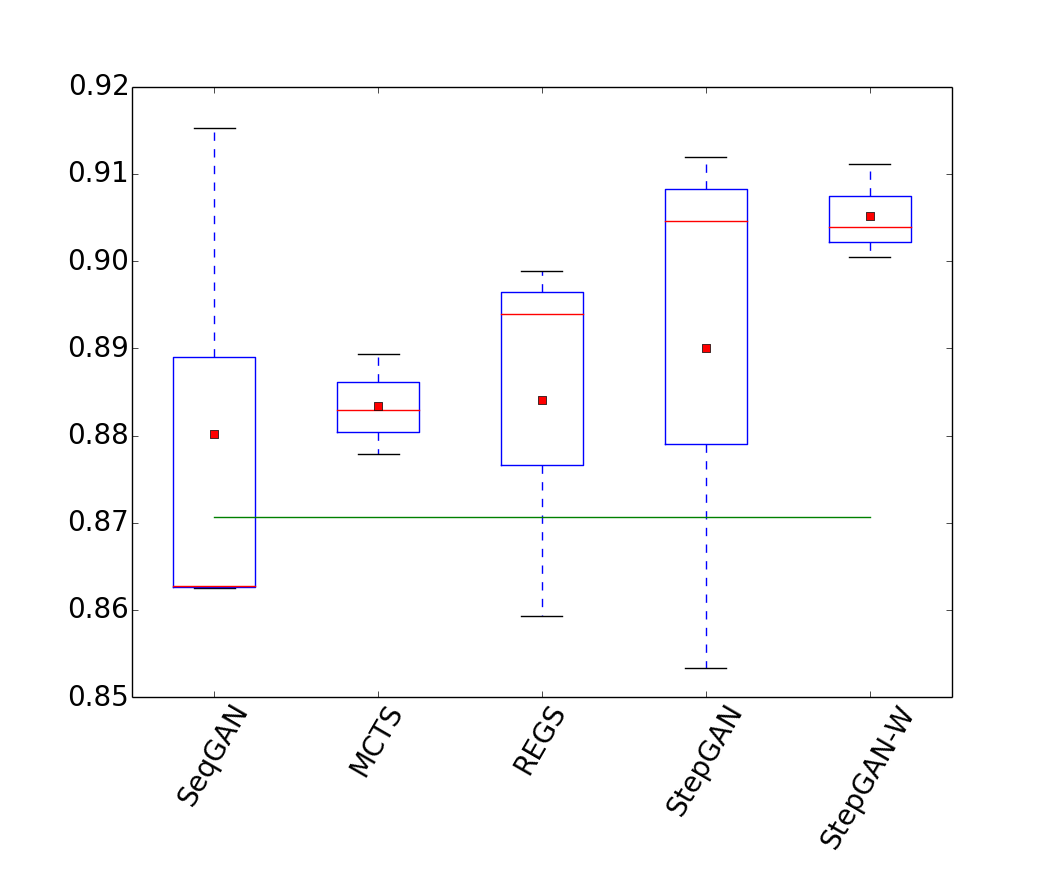}
\caption{The box plot of results obtained by different random parameter initialization. 
Each box represents the results of   each model in Table~\ref{tab:pretrain-count} trained with different random initialization. 
Green line represents the precision result of MLE.}
\label{fig:syn-best-params}
\end{figure}

In Figure~\ref{fig:syn-best-params}, we show the variance of the performance of the GAN-based approaches with different random parameter initialization.
Each box represents the results of each model in Table~\ref{tab:pretrain-count} trained with different random initialization, and the green line represents the precision result of MLE.
The results show that StepGAN-W not only achieves better performance than other approaches, it is also more stable with different parameter initialization.

{\bf Forward and Inversed KL-Divergence.} Comparing to MLE, we observe that GANs increase FKLD (except SeqGAN) and decrease IKLD.
This is reasonable because GANs do not minimize FKLD as MLE, but minimize other distance metrics.
Because minimizing FKLD will cause the $P_G$ unbounded at where $P_R$ is almost zero (Equation~(\ref{eq:KLD})), and vise versa, we consider that FKLD and IKLD are equally important to measure true distance between two distributions.
Hence, we add them together and get a score FKLD+IKLD to check which model is overall good.
The SeqGAN and previous stepwise evaluation methods improve FKLD+IKLD score by $0.009\sim 0.071$; our approaches StepGAN and StepGAN-W respectively improve this divergence score by $0.162$ and $0.31$, much better than previous methods.

The results show that GANs fine-tune a pretrained model that has converged to reach a better performance and reduce IKLD, and the precision scores are thus higher than the pretrained model.
Furthermore, the stepwise evaluation methods can improve or maintain the model to generate as much as possible correct answers by showing comparable recall scores. 
Among them, StepGAN and StepGAN-W can reach better precision with little penalty of recall, and have more balanced results of FKLD+IKLD scores.

\section{Experiments: Chit-chat Dialogue Generation}
\label{sec:chat}

We trained chit-chat dialogue generation on OpenSubtitles\cite{Tiedemann:RANLP5}, a collection of movies subtitles.  
After pretraining the generator by MLE, we fine-tuned the model for 1-epoch by three different stepwise evaluation methods: vanilla SeqGAN, REGS, and StepGAN
We choose to compare vanilla SeqGAN and REGS because  they have been reported helpful on dialogue generation task~\cite{li2017adversarial}. We do not have the results of MCTS here because it costs too much computation which is not tractable on our machine.

\subsection{Experimental Setup}

\begin{figure}[t!]
    \centering
    \includegraphics[width=\linewidth]{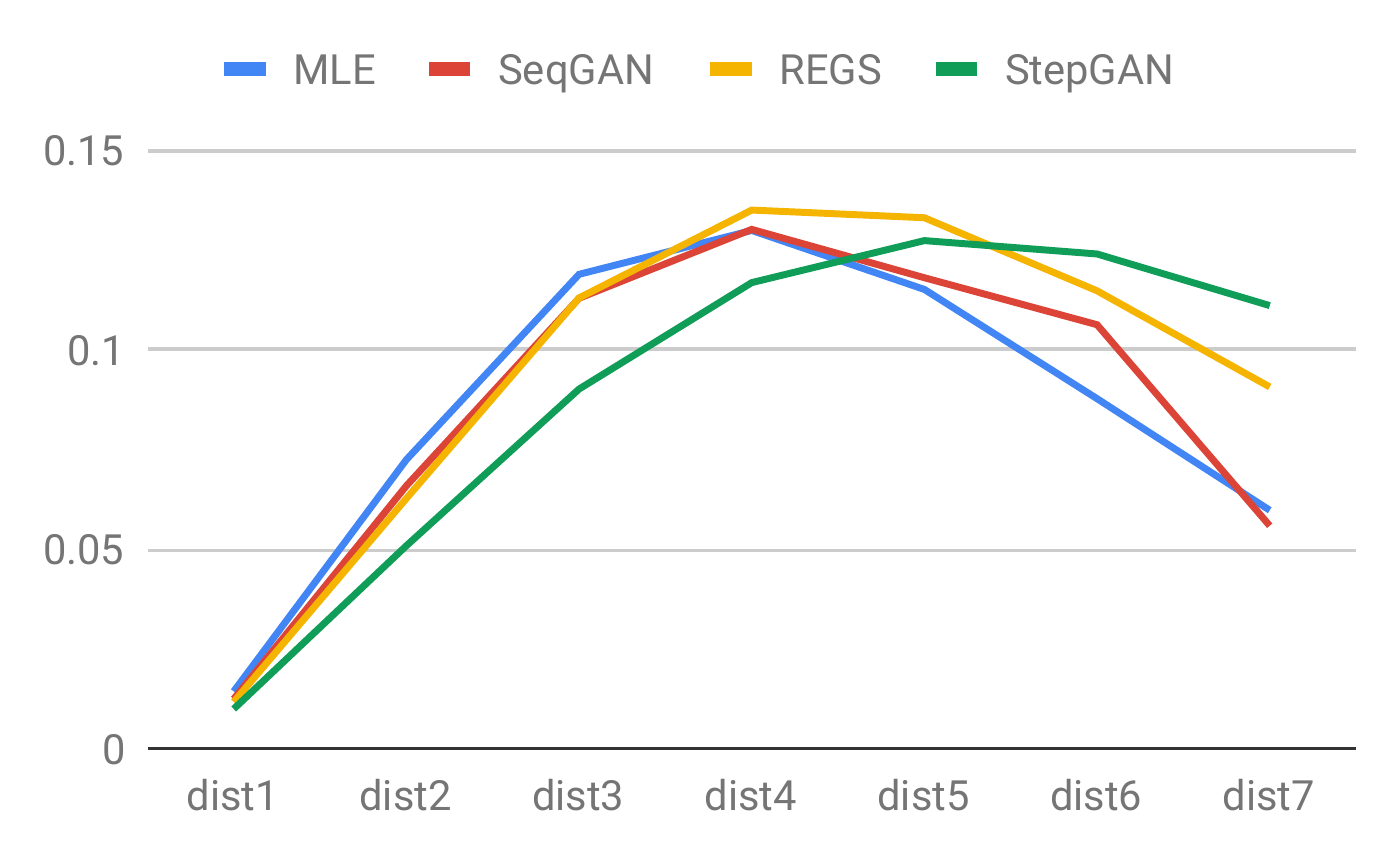}
    \caption{\textcolor{black}{The results of dist-N for chit-chat dialogue generation.}}
    \label{fig:dist-n}
\end{figure}

We trained chit-chat dialogue generation on OpenSubtitles\citep{Tiedemann:RANLP5} and used the top $4000$ most frequent words as the vocabulary.
After pretraining the generator by MLE, we trained the model for 1-epoch by four different stepwise evaluation methods: vanilla SeqGAN, REGS, and StepGAN.
All the discriminators were pre-trained on real data and generated data sampled from the pre-trained generator. 
To tune the parameters, grid search was used with optimization operation=\{SGD, Adam, RMSProp\}, learning rate=\{1e-1,1e-2,1e-3,1e-4\}, discriminator iteration step=\{1,5\}, and batch size=\{32,64\}.
All the generators and discriminators are 1 GRU layer with 512 dimension, which is an acceptable number of parameters on sentence generation~\citep{press2017language,rajeswar2017adversarial}.

\begin{figure}[t!]
\centering
\subfloat[positive example.]{{
    \includegraphics[trim={0 0 2cm 0},clip,width=.45\linewidth]{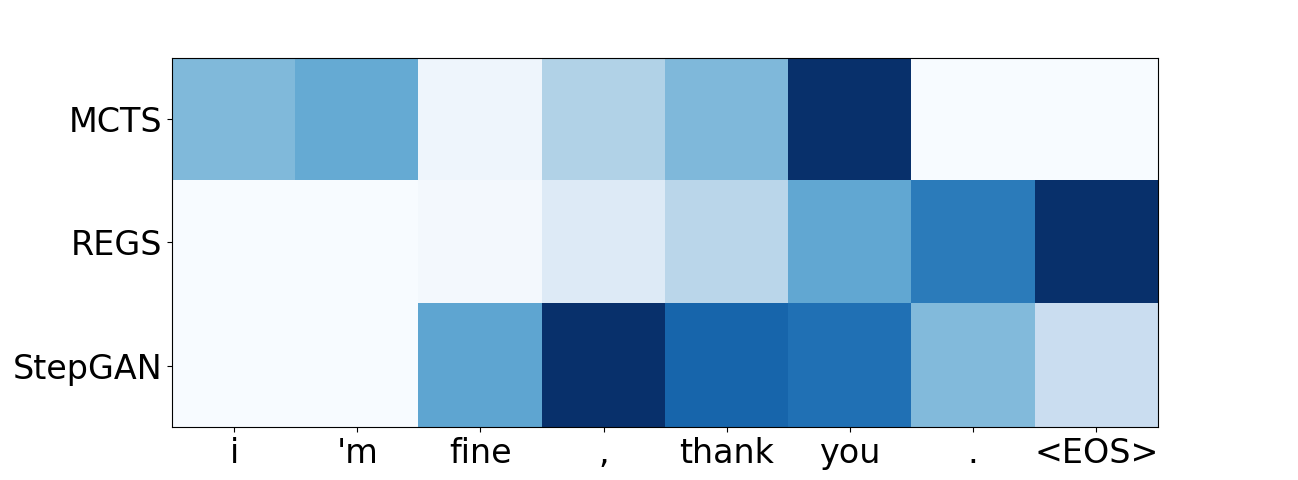}}
    \label{fig:var-1}}\hfill
\subfloat[negative example.]{{
    \includegraphics[trim={0 0 2cm 0},clip,width=.45\linewidth]{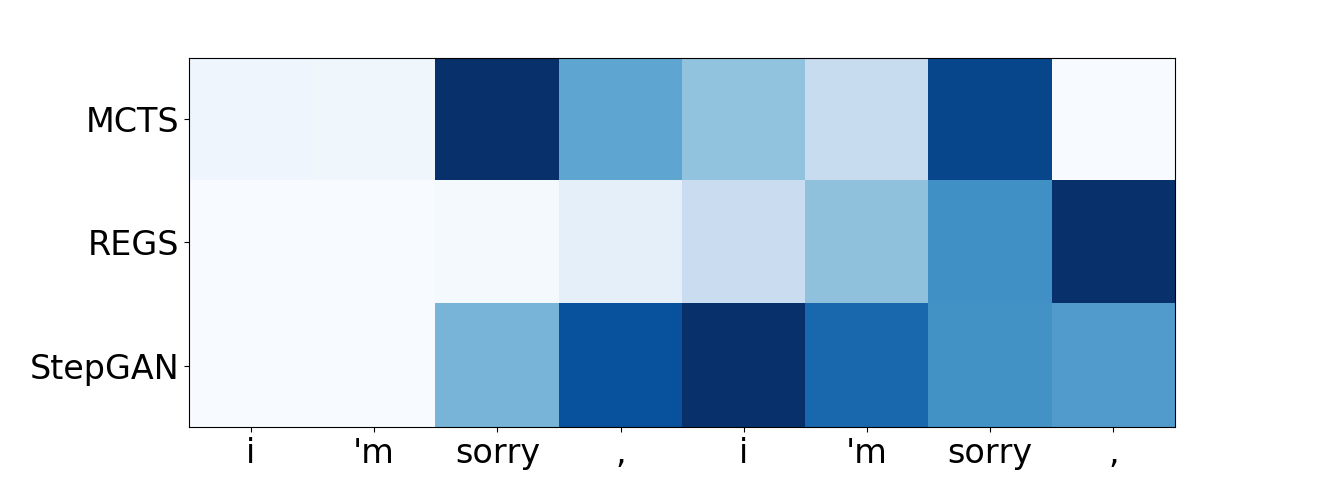}}
    \label{fig:var-2}}\\
\subfloat[positive example.]{{
    \includegraphics[trim={0 0 2cm 0},clip,width=.45\linewidth]{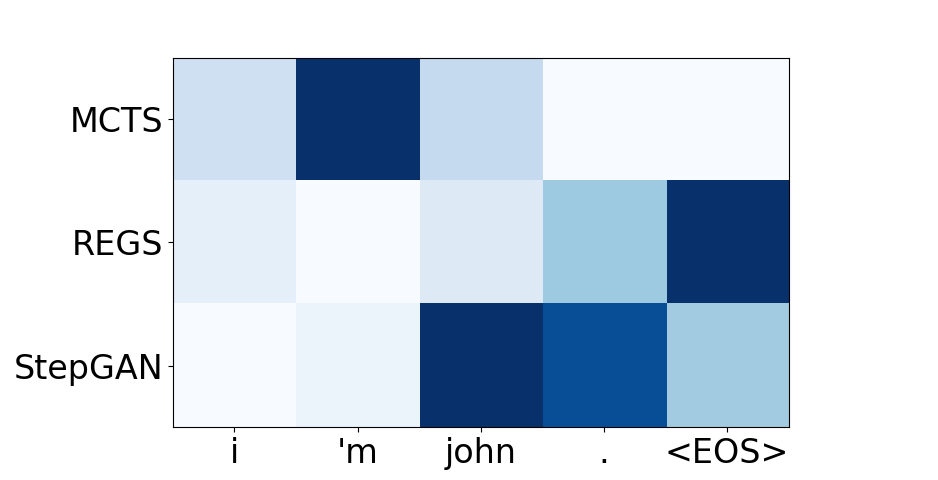}}
    \label{fig:var-3}}\hfill
\subfloat[negative example.]{{
    \includegraphics[trim={0 0 2cm 0},clip,width=.45\linewidth]{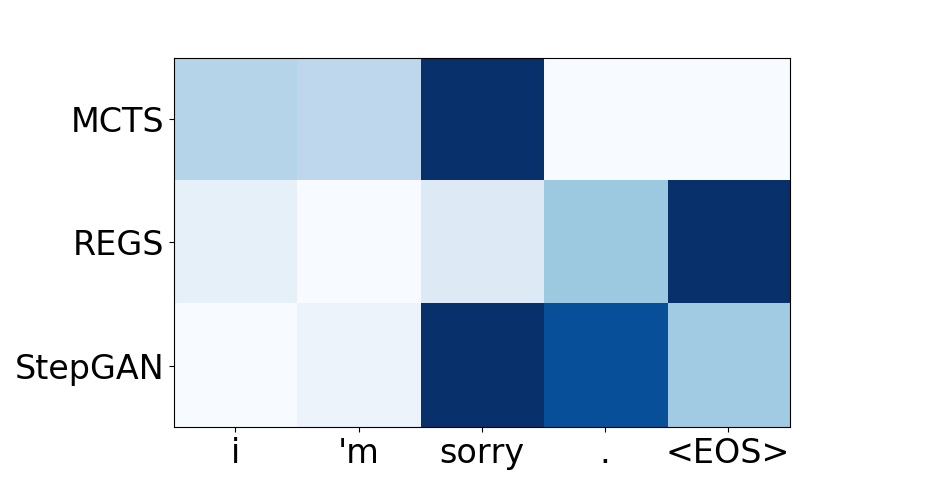}}
    \label{fig:var-4}}\\
\caption{The variance of learned state-action value during training. The higher variance is represented by darker color. (a)(b) are given \emph{``how are you ?''} as input, and (c)(d) are given \emph{``what 's your name ?''} as input.}
\label{fig:Var}
\end{figure}

\begin{figure}[t!]
\centering
    \includegraphics[width=\linewidth]{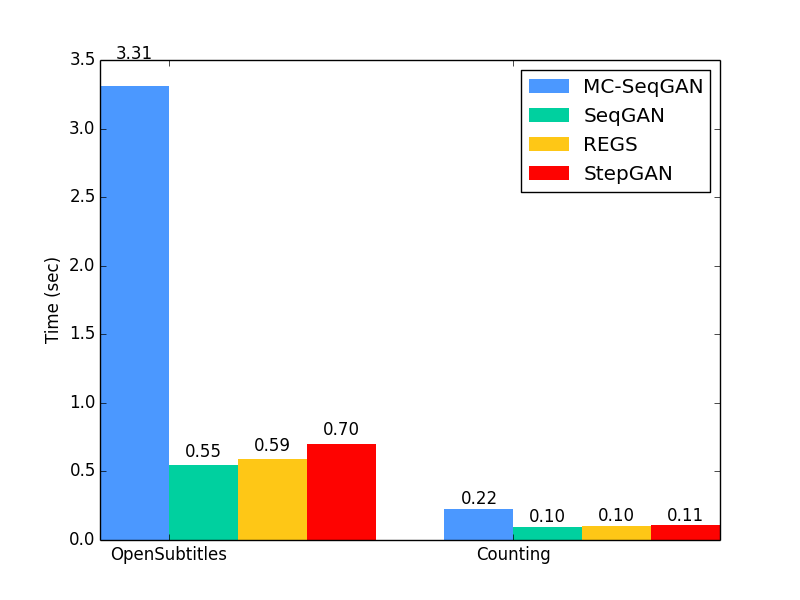}
\caption{The average computation times for  one training iteration (i.e., one mini-batch) of chit-chat dialogue generation (OpenSubtitles) and synthetic experiment (Counting).}
\label{fig:times}
\end{figure}

\subsection{Evaluation Metrics} 
Currently, chit-chat dialogue generation still lack general rules for evaluation~\cite{liu2016not}, therefore we use several different metrics to evaluate the generated responses. 

{\bf BLEU.} BLEU score~\cite{papineni2002bleu} is an automatic evaluation metric that counts the appearance of n-grams. Although BLEU has been reported not consistent with the human evaluation~\cite{liu2016not} on dialogue, we reported it here as a reference.

{\bf Coherence Human Score (CoHS) and Semantics Human Score (SHS).} We invited 5 judges to evaluate 200 examples.
Because coherence is naturally binary (yes or no) and semantic is continuous, we asked judges to measure them by 0-1 test and ranking respectively.
For {\bf CoHS}, judges were asked to measure if each response is coherent to the input sentence (0 or 1); for {\bf SHS}, they were asked to rank the information abundance of the given responses.
The ranking is then normalized to a score ranging from 0 to 1.
Both CoHS and SHS are the higher the better.

{\bf LEN and GErr.} {\bf LEN} stands for lenth and is the average of length of generated responses; {\bf GErr} stands for grammar error, and we measured it by an open source software \texttt{https://}
\texttt{github.com/languagetool-org/languagetool} and calculated the number of violations of grammar rules.

{\bf General.} This metric counts the number of generated general responses.
Here we define the top $10\%$ responses generated from the model learned by MLE as general responses.
The top two general responses are ``I don't know'' and ``I'm sorry''.
However, if a ground truth response is one of the top $10\%$ responses, it will not be counted as a general response. 

\textcolor{black}{{\bf Dist-N.} This metric counts the number of distinct n-grams and divides it by the total number of words~\cite{li2016diversity}.}

\subsection{Results}
The results of our proposed evaluation metrics are listed in Table~\ref{tab:main-dialog}. Generally, we found that on CoHS, MLE gets the highest score; on BLEU, SeqGAN is the highest; on metrics related to the quality of generated sentences (i.e., SHS, LEN, GErr, and General), StepGAN consistently gets the best scores.

We observed that all the three types of GANs can generate sentences with better semantic scores (SHS), longer lengths (LEN) and less general responses (General), but they do not equally lower GErr. 
Table~\ref{tab:main-dialog} shows that among them, StepGAN is the most beneficial one, which yielded the highest semantic human score (SHS), longest generated sentences (LEN), the least grammatical errors (GErr) and the least general responses (General). 
This shows that the stepwise evaluation methods surely impact the results.  
The assumption that MLE easily results in general responses~\cite{li2016diversity} can also be verified by the observation.
Even MLE has the highest CoHS, it has relatively low SHS and LEN.
This is probably because CoHS is more related to conditioning rather than sequence modeling. 

\textcolor{black}{The Dist-N metrics are plotted in Figure~\ref{fig:dist-n}. Compared to MLE, GANs do not have better diversity when N is below 4-gram but show diversity when N is above 5-gram. Specifically, when N is larger, StepGAN shows much better performance than others. This demonstrates the ability of StepGAN to strengthen the benefits of GANs.}

In Table~\ref{tab:example}, we present some generated examples of our trained dialogue generative models.

\begin{table*}[t]
\begin{center}
\resizebox{.75\linewidth}{!}{
\begin{tabular}{|l|ccc|ccc|}\hline
\multirow{2}{*}{\bf Model-InferMethod} &\multicolumn{3}{c|}{\bf CoHS~(\%)} &\multicolumn{3}{c|}{\bf SHS~(\%)}\\\cline{2-7}
& \bf argmax & \bf BS & \bf MMI & \bf argmax & \bf BS & \bf MMI\\\hline
{\bf MLE} & 50.52 & \bf 58.86 & 53.68 & 42.6 & 8.90 & 10.0\\
{\bf SeqGAN} & 46.25 & 56.87 & 54.34 & 56.4 & 14.3 & 15.3\\
{\bf MaliGAN} & 36.31 & 48.44 & 40.85 & 23.4 & 8.4 & 9.2\\
{\bf REGS} & 48.43 & 55.21 & 50.92 & 59.5 & 13.6 & 21.1\\
{\bf StepGAN-W} & 44.96 & 53.47 & 50.88 &\bf 60.4 & 10.0 & 18.3\\\hline
\end{tabular}
}
\caption{The results of human evaluation for chit-chat dialogue generation with different inference methods: argmax, beam search(BS) and maximum mutual information (MMI).}
\label{human-eval}
\end{center}
\end{table*}

\subsection{CoHS and SHS Results}

\textcolor{black}{To survey the effectiveness of different inference methods,} we asked 5 judges to evaluate 200 random selected examples. Each example consists of an input and 15 generated outputs by five different training algorithms and three inference methods. The training algorithms include MLE, SeqGAN, MaliGAN, REGS, and StepGAN; the inference methods include argmax, beam search, and MMI~\cite{li2016diversity}.

\textcolor{black}{As shown in Table~\ref{human-eval}, MLE with beam search achieves the best CoHS, while StepGAN-W with argmax achieves the best SHS. In terms of training methods, this phenomenon is consistent with the results in Table~\ref{tab:main-dialog}.
For inference methods, although MMI has better SHSs and weaker CoHSs than beam search, both beam search and MMI lead to higher CoHSs but lower SHSs.
This indicates that these inference methods tend to assure coherence instead of informativeness, thus eliminating the benefits of GANs.
Therefore argmax is suggested to be used for sequence GANs.}

\subsection{Analysis}

\begin{figure}[t!]
\centering
    \includegraphics[width=\linewidth]{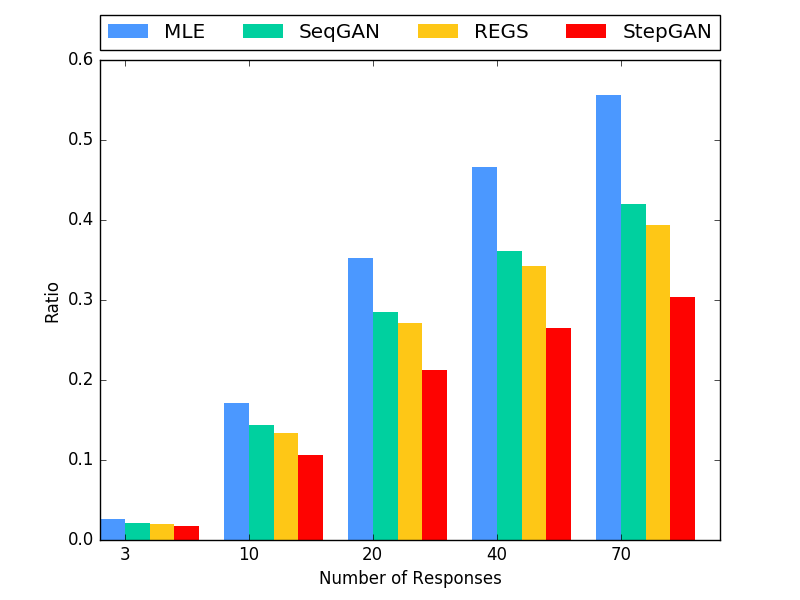}
    \caption[font=small]{The proportion of generated responses that have appeared in the training data.}
\label{fig:value}
\end{figure}

\subsubsection{discriminator outputs}

To verify if the proposed stepwise evaluation method can estimate the state-action values, we analyzed the outputs of discriminators. 
Empirically, given the same response, the discriminator scores $D$ change frequently between the training iterations. 
This makes the values $\hat{Q}(s_t,y_{t})$ from $D$ unstable and thus difficult to be analyzed. 
We proposed to measure \emph{the variance at each generation step throughout the training iterations}. 
During training, $\hat{Q}(s_t,y_{t})$ oscillates  according to the performance of generator at that time. 
If at the generation step $t$, $\hat{Q}(s_t,y_{t})$ oscillates violently, the word generated at step $t$ is crucial to determine if a response is fake or real. 

Four examples are shown in Figure~\ref{fig:Var}. 
Figure~\ref{fig:var-1} and~\ref{fig:var-2} are respectively true response and wrong response given input sentence \emph{``how are you ?''}; Figure~\ref{fig:var-3} and~\ref{fig:var-4} are true and wrong responses given input sentence \emph{``what 's your name ?''}. 
The depth of the color is proportional to the variance at each generation step for MCTS\footnote{In Figure~\ref{fig:Var}, the MCTS results are calculated by using MCTS on SeqGAN model.}, REGS and StepGAN.

We can observe that the variances of StepGAN and MCTS are similar, and the generation steps with large variance, that is, the critical words in the response, are consistent with human intuition. 
For example, when given \emph{``what 's your name ?''}, they focus on the bold parts of \emph{``i \textbf{'m john} .''} in Figure~\ref{fig:var-3} and \emph{``i 'm \textbf{sorry} .''} in Figure~\ref{fig:var-4}.
\textcolor{black}{People cannot identify if ``i 'm'' is good or bad, but can identify if the sentence is good or bad with more information.}
Besides, we found that the variances of StepGAN are even more reasonable for human than MCTS (e.g., (a) i 'm {\bf fine}, ...... (c) i 'm {\bf john}). 
The most possible reason is that the variances of StepGAN are directly an expectation of all the training examples; the variances of MCTS are average of $I=5$ roll-out paths and highly depend on the number $I$.
\textcolor{black}{Figure~\ref{fig:Var}~(a-d) also shows that REGS puts emphases on the last few steps, and the critical words do not meet human's intuition.}

\subsubsection{computation}

As demonstrated by Figure~\ref{fig:times}, StepGAN is much more computational efficiency than MCTS.
The computational consumption of MCTS depends on the number of roll-out paths which is set to $5$ in Figure~\ref{fig:times}.
\textcolor{black}{In synthetic task, SeqGAN, REGS, and StepGAN are twice faster than MCTS; in dialogue modeling, they are about five times faster than MCTS.
For longer sequence length and larger number of roll-out paths, the time consumption of MCTS grows.
MCTS is therefore intractable for  large datasets.}

\subsubsection{general responses}

To see how the GANs learn from the training data, we first sorted the training data into input-outputs pairs -- each input was paired with all the appeared outputs in training data. We then used the models trained by MLE, SeqGAN, REGS and StepGAN to generate responses, and counted how many responses are in the training data. The ratio of the number of appeared responses to the number of inputs is shown in Figure~\ref{fig:value}.
\textcolor{black}{The results verify that GANs learn more creative responses~\citep{yu2017seqgan} that are not in training data. StepGAN especially generates the least non-creative ones.}

\section{Conclusion}
This paper verifies that stepwise evaluation methods have notable impact on conditional sequence generation, and proposes a novel stepwise evaluation method -- StepGAN that can directly estimate state-action value by discriminator.
In experiments, we show that StepGAN can help conditional sequence generation. 
Compared to MCTS, it is computational efficient; compared to vanilla SeqGAN and REGS, it is more accurate on synthetic experiment and generates sentences with higher quality in dialogue generation.

\begin{table*}[t]
\begin{center}
\begin{tabular}{r|l}\hline
\bf input & i 'll send ya some more money .\\\hline
\bf ground-truth & that man hitting you ?\\\hline
\bf MLE & that 's a good idea .\\
\bf SeqGAN & that 's a good idea .\\
\bf REGS & that 's a nice car , huh ?\\
\bf StepGAN-W & you 're a good man , you know that ?\\
\hline\hline
\bf input & i was so stupid to think that anything could ever change you .\\\hline
\bf ground-truth & you don 't want a child , so you agree not to be a part of that child 's life . \\\hline
\bf MLE & i 'm sorry .\\
\bf SeqGAN & i 'm sorry , i 'm sorry .\\
\bf REGS & i don 't want to be here anymore .\\
\bf StepGAN-W & you 're gonna be all right .\\

\hline\hline
\bf input & fiona , my love , we gonna kiss away for our happily ever after .\\\hline
\bf ground-truth & now kiss me ! \\\hline
\bf MLE & oh , my god .\\
\bf SeqGAN & oh , my god .\\
\bf REGS & oh , my god .\\
\bf StepGAN-W & oh , i love you .\\

\hline\hline
\bf input & i had some girlfriends too , but all they wanted from me was weed and shit .\\\hline
\bf ground-truth & shit , my grandmother used to say , " what 's better ? \\\hline
\bf MLE & but , i don 't know what to do .\\
\bf SeqGAN & i am sorry , but i 'm not a doctor .\\
\bf REGS & i was just kidding .\\
\bf StepGAN-W & what 's wrong with you , man ?\\

\hline\hline
\bf input & tell me how this ends .\\\hline
\bf ground-truth & the stage is almost set . \\\hline
\bf MLE & i 'm not going to be able to get you .\\
\bf SeqGAN & i 'm not going to be able to get you .\\
\bf REGS & i 'm not going to be able to be a baby .\\
\bf StepGAN-W & fine , you 're good .\\

\hline\hline
\bf input & we gave it a thousand times more power .\\\hline
\bf ground-truth & we tried to send something from new mexico to a twin machine at a lab in new york city . \\\hline
\bf MLE & i 'm sorry .\\
\bf SeqGAN & i 'm not a doctor .\\
\bf REGS & i don 't know what to do .\\
\bf StepGAN-W & you were right .\\

\hline\hline
\bf Input & i feel like a paris princess .\\\hline
\bf ground-truth & what yall doing in that tub ? \\\hline
\bf MLE & i know it .\\
\bf SeqGAN & i know that you 're not there .\\
\bf REGS & i can 't see it .\\
\bf StepGAN & you 're not a child anymore .\\

\hline\hline
\bf Input & i 'm tired of all your questions .\\\hline
\bf ground-truth & i 'm not here to teach you everything about everything in the world . \\\hline
\bf MLE & i 'm gonna have to ask you to come .\\
\bf SeqGAN & i 'm gonna call you .\\
\bf REGS & i 'm going to have to ask you to come in\\
\bf StepGAN & i 'm going to be a friend of you .\\


\hline\hline
\bf Input & he 's one of your own .\\\hline
\bf ground-truth & hang on . \\\hline
\bf MLE & what ?\\
\bf SeqGAN & what are you doing here ?\\
\bf REGS & he 's not my father\\
\bf StepGAN & what 's your name , man ?\\

\hline\hline
\bf Input & what dance will you have ?\\\hline
\bf ground-truth & a coranto ? \\\hline
\bf MLE & i don 't know .\\
\bf SeqGAN & i 'm not sure that i 'm not going to .\\
\bf REGS & i don 't know .\\
\bf StepGAN & good night , i 'll tell you what 's going on .\\





\hline\hline
\bf Input & i was just gonna tell you that your fly is down .\\\hline
\bf ground-truth & manager ! \\\hline
\bf MLE & you 're not gonna die .\\
\bf SeqGAN & you are not a liar .\\
\bf REGS & you 're not gonna die , but i don 't have to .\\
\bf StepGAN & you 're gonna have to help me out .\\

\hline\hline
\bf Input & will you come on now , please ?\\\hline
\bf ground-truth & come on . \\\hline
\bf MLE & i 'm not going to let you go .\\
\bf SeqGAN & i can 't hear you .\\
\bf REGS & i can 't believe it .\\
\bf StepGAN & what 's this all about ?\\


\end{tabular}
\caption{Examples of generated responses.}
\label{tab:example}
\end{center}
\end{table*}

\ifCLASSOPTIONcaptionsoff
  \newpage
\fi

\bibliography{stepgan}
\bibliographystyle{IEEEtran}

\end{document}